\documentclass[11pt, oneside]{article}
\usepackage[english]{babel}
\usepackage{geometry}  
\usepackage{graphicx}      
\usepackage{tabularx}      
\usepackage{amssymb}
\usepackage{amsmath}
\usepackage{eulervm}
\usepackage{xltabular}
\usepackage[hyphens]{url}
\usepackage[hidelinks]{hyperref}
\hypersetup{breaklinks=true}
\usepackage[backend=biber,style=nature, doi=false,isbn=false, date=year, url=false, autopunct=true]{biblatex}
\makeatletter
\renewbibmacro*{textcite}{%
  \iffieldequals{namehash}{\cbx@lasthash}
    {\usebibmacro{cite:comp}}
    {\usebibmacro{cite:dump}%
     \ifbool{cbx:parens}
       {\bibclosebracket\global\boolfalse{cbx:parens}}
       {}%
     \iffirstcitekey
       {}
       {\textcitedelim}%
     \usebibmacro{cite:init}%
     \ifnameundef{labelname}
       {\printfield[citetitle]{labeltitle}}
       {\printnames{labelname}}%
     \addspace
     \ifnumequal{\value{citecount}}{1}
       {\usebibmacro{prenote}}
       {}%
     \mkbibsuperscript{\usebibmacro{cite:comp}}%
     \stepcounter{textcitecount}%
     \savefield{namehash}{\cbx@lasthash}}}
\makeatother

\usepackage{csquotes}
\usepackage{cleveref}
\usepackage{siunitx}
\usepackage{geometry}
\usepackage{multirow}
\usepackage[dvipsnames,table,xcdraw]{xcolor}
\definecolor{tumblue}{rgb}{0.235,0.059,0.439}

\usepackage{xspace}
\usepackage{caption}
\usepackage{subcaption}
\usepackage{authblk}

\usepackage{marvosym}

\usepackage{sectsty} 

\sectionfont{\color{tumblue}\selectfont\sffamily}
\subsectionfont{\color{tumblue}\selectfont\sffamily}
\subsubsectionfont{\color{tumblue}\selectfont\sffamily}
\paragraphfont{\selectfont\sffamily}
\subparagraphfont{\color{gray}\selectfont\sffamily}

\definecolor{halfgray}{gray}{0.55}

\IfFileExists{./fonts/Heliotrope-Bold.otf}{
    \usepackage{fontspec}
    \setmainfont{Heliotrope}[Extension = .otf, UprightFont = *-Regular, ItalicFont = *-Italic, BoldFont=*-Bold, Path = ./fonts/]
}{
    \usepackage{biolinum}
    
}

\usepackage[labelfont=bf]{caption}

 \usepackage{microtype}
 \usepackage{fontawesome}

\definecolor{linkcolor}{RGB}{0, 102, 204} 

\renewcommand{\thefootnote}{\roman{footnote}}

\usepackage{credits}
\usepackage{orcidlink}
\usepackage{multirow}
\usepackage{amssymb}
\usepackage{pifont}
\usepackage{booktabs}
\usepackage{tcolorbox}
\usepackage{listings}
\usepackage{tabularx}
\usepackage{minted}
\usepackage{longtable}
\usepackage[utf8]{inputenc}
\usepackage[section]{placeins}
\usepackage{cleveref}
\usepackage{svg}
\usepackage{caption}
\usepackage{subcaption}

\newcommand{\pxrd}{pXRD}
\newcommand{\dos}{DOS}

\title{MatBind: A Shared Embedding Space for Multimodal Materials Characterization}
\credit{Le Yang}{1,1,1,0,1,1,0,0,1,0,1,1,1,1}
\credit{Anoop~K.~Chandran}{1,1,1,0,1,1,0,0,1,0,1,1,0,1}
\credit{Jona~Oestreicher}{1,1,1,0,1,1,0,0,1,0,1,1,0,1}
\credit{Evgenii~Sovetkin}{1,1,1,0,1,1,0,0,1,0,1,1,0,1}
\credit{Adrian~Mirza}{1,1,1,0,1,1,0,0,1,0,1,1,0,1}
\credit{Sebastien~Bompas~}{1,1,1,0,1,1,0,0,1,0,1,1,0,1}
\credit{Bashir~Kazimi~}{1,1,1,0,1,1,0,0,1,0,1,1,0,1}
\credit{Pascal~Friederich~}{1,1,1,0,1,1,0,0,1,0,1,1,0,1}
\credit{Stefan~Kesselheim}{1,1,1,0,1,1,0,0,1,0,1,1,0,1}
\credit{Kevin Maik Jablonka}{1,1,1,1,1,1,0,1,1,1,1,1,0,1}
\credit{Stefan~Sandfeld~}{1,0,1,1,1,1,1,1,0,1,1,1,0,1}

\newcommand{\corrauth}{$^{*}$}

\author[1]{Le~Yang~\orcidlink{0009-0002-0944-9675}}
\author[2]{Anoop~K.~Chandran\orcidlink{0000-0001-9078-0132}}
\author[3]{Jona~Östreicher}
\author[2]{Evgenii~Sovetkin~\orcidlink{0000-0002-1951-4971}}
\author[4,6]{Adrian~Mirza~\orcidlink{0000-0003-4033-4235}}
\author[1]{Sebastien~Bompas~\orcidlink{0009-0003-6818-0861}}
\author[1]{Bashir~Kazimi~\orcidlink{0000-0003-1802-7511}}
\author[3]{Pascal~Friederich~\orcidlink{0000-0003-4465-1465}}
\author[2,7]{Stefan~Kesselheim~\orcidlink{0000-0003-0940-5752}}
\author[6,8,9]{Kevin~Maik~Jablonka~\orcidlink{0000-0003-4894-4660}\corrauth}
\author[1,5]{Stefan~Sandfeld~\orcidlink{0000-0001-9560-4728}\corrauth}

\affil[1]{Institute for Advanced Simulations (IAS-9),
Forschungszentrum Jülich GmbH,
52425 Jülich, Germany }
\affil[2]{Jülich Supercomputing Centre,
Forschungszentrum Jülich GmbH,
52425 Jülich, Germany}
\affil[3]{Institute of Nanotechnology, Karlsruhe Institute of Technology, 76131 Karlsruhe, Germany}
\affil[4]{Helmholtz-Zentrum Berlin für Materialien und Energie GmbH, Hahn-Meitner-Platz 1, 14109, Berlin, Germany}
\affil[5]{Faculty 5 – Georesources and Materials Engineering,
RWTH Aachen University,
Aachen 52056, Germany }
\affil[6]{Helmholtz Institute for Polymers in Energy Applications Jena (Jena), Lessingstraße 12--14, 07743 Jena, Germany}
\affil[7]{1. Phys Inst, University of Cologne, Zülpicher Str. 77, 50937, Köln, Germany}
\affil[8]{Laboratory of Organic and Macromolecular Chemistry, Friedrich Schiller University Jena, Humboldstr. 10, 07743 Jena, Germany}
\affil[9]{Center for Energy and Environmental Chemistry Jena, Friedrich Schiller University Jena, Philosophenweg 7, 07743 Jena, Germany}
\begin{document}
\maketitle
\begingroup
\renewcommand{\thefootnote}{}
\footnotetext{$^{*}$Corresponding authors: \texttt{mail@kjablonka.com, s.sandfeld@fz-juelich.de}}
\endgroup
\begin{abstract}
Fully characterizing a crystalline material requires integrating heterogeneous data sources --- atomic structures, diffraction 
patterns, electronic density of states, and natural language --- each of which captures a different facet of the same physical 
object. 
In practice, however, these modalities are stored and analyzed in isolation, making it difficult to relate or query 
materials across representational boundaries. 
We present \textbf{MatBind}, a contrastive learning framework that aligns four materials modalities --- crystal structure, powder X-ray diffraction (\pxrd) simulated from structures, density of states (\dos), and text --- into a unified embedding space using crystal structure as the central physical anchor. 
The framework induces alignment between modalities never explicitly paired during training, enabling emergent zero-shot cross-modal retrieval as a direct consequence of the shared representation. The learned embedding space organizes materials according to physically meaningful properties without explicit supervision, and retrieval performance improves systematically when modalities are combined at query time. These results demonstrate that treating heterogeneous materials data as complementary projections of a single physical reality, rather than as isolated data sources, is not a practical choice but is consistent with the underlying physics.
\end{abstract}

\section{Introduction}

Materials data is inherently multimodal. In routine practice, researchers move between atomistic structure models, powder X-ray diffraction (\pxrd) patterns, electronic density of states (\dos), and textual descriptions from the literature to determine what a material is and how it behaves. These representations are not independent views of a material, but complementary constraints on the same underlying system: structure governs diffraction and electronic response, while text often encodes composition, symmetry, and prior domain knowledge. A framework that can relate these modalities within a common representation would therefore more closely reflect how materials are actually analyzed and identified.

In practice, however, this inherent connection between different modalities is rarely exploited. Experimental measurements, computational simulations, and textual knowledge live in separate databases and are analyzed using modality-specific tools. As a result, it is difficult to relate, compare, or query materials across heterogeneous representations: a measured diffraction pattern cannot easily be used to retrieve materials with similar electronic properties, and a target electronic signature cannot be directly linked to known experimental observations. This fragmentation limits data-driven materials discovery and forces researchers to artificially separate information that is physically connected.

Recent advances in contrastive learning offer a route to overcoming this fragmentation. Since the introduction of SimCLR~\cite{chen2020simple}, contrastive objectives have proven effective at learning structured, transferable representations by bringing different views of the same entity closer in a shared embedding space. CLIP~\cite{radford2021learning} extended this to cross-modal alignment of images and text, and ImageBind~\cite{girdhar2023imagebind} generalized it further to seven distinct modalities by using a single ``anchor'' modality to induce alignment across all pairs. By requiring alignment only between each modality and the anchor, the framework naturally accommodates missing modalities. In materials science, contrastive learning has begun to be applied to crystal structures~\cite{koker2022graph}, to aligning structures with \pxrd\ 
patterns~\cite{xu2025kan}, and to connecting crystal graphs with textual descriptions~\cite{park2025contrastive,suzuki2025bridging}. However, these efforts have addressed at most two modalities at a time and do not address the broader challenge of unified multimodal alignment across the full range of experimentally and computationally accessible representations. 

\begin{figure}[h]
    \centering
    \includegraphics[width=\linewidth]{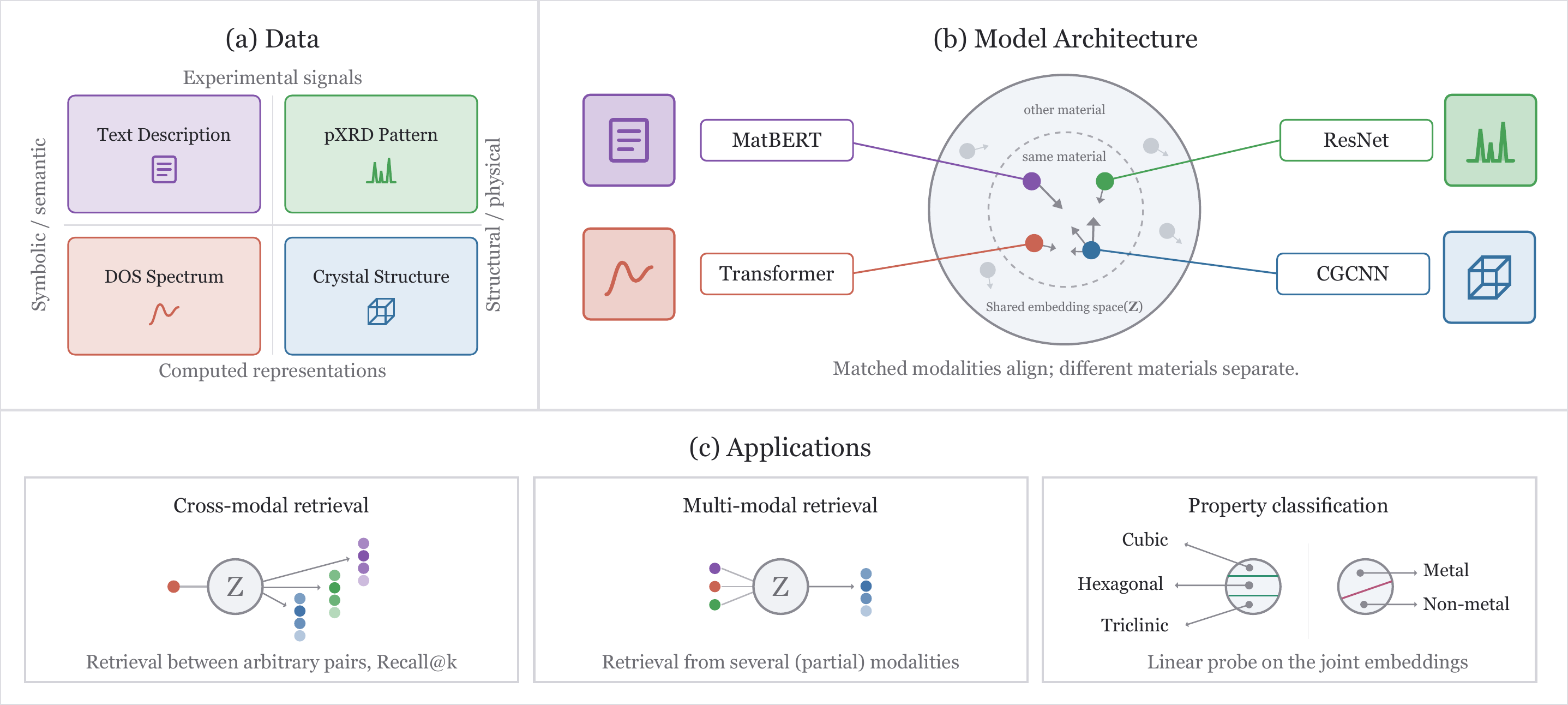}
    \caption{\textbf{MatBind aligns diverse materials modalities into a unified embedding space for retrieval and property prediction.} (a) Four material modalities are organized along symbolic-structure and experiment-simulation axes: \dos, text descriptions, \pxrd\ patterns, and crystal structures. (b) Each modality is processed by a dedicated encoder that projects inputs into a unified joint embedding space. (c) The framework can be applied to different tasks, such as cross-modal retrieval between arbitrary modality pairs, multimodal retrieval from partial or combined inputs, and material property classification via a linear probe on joint embeddings.
    }
    \label{fig:overview}
\end{figure}

Here we introduce \textbf{MatBind}, a contrastive learning framework that aligns four key materials modalities --- crystal structure, \pxrd, \dos, and text --- into a single embedding space (Figure~\ref{fig:overview}). 
Inspired by the anchor-based training strategy of ImageBind, we adopt the crystal structure as the central modality in our setting and train pairwise contrastive objectives between the structure and each auxiliary modality. This yields a shared representation in which all four modalities become mutually comparable, including modality pairs that are never explicitly aligned during training. Throughout this paper, we denote the bidirectional cross-modal retrieval task (or, equivalently, the unordered modality pair) between modalities A and B by A$\leftrightarrow$B. We demonstrate that:

\begin{itemize}
    \item the learned embedding space organizes materials according to physically meaningful properties (e.g., band gap and crystal system) without explicit supervision on these  quantities;
    \item cross-modal retrieval between directly trained modality pairs achieves high recall, with crystal structure 
    $\leftrightarrow$ text and crystal structure $\leftrightarrow$ \dos\ approaching near-perfect performance;
    \item emergent zero-shot retrieval between modality pairs never explicitly aligned during training in some cases surpasses directly trained pairs; most notably, the emergent \dos$\leftrightarrow$text link outperforms the directly trained crystal structure$\leftrightarrow$\pxrd\ pair;
    \item querying the embedding space with multiple modalities simultaneously resolves ambiguities that single modality, like \pxrd\ alone cannot, yielding measurable gains in structure identification accuracy.
\end{itemize}

In this work, we focus on a controlled benchmark setting based on Materials Project structures. The pXRD patterns are simulated and the text modality consists of automatically generated crystal-structure descriptions. This setting allows us to isolate the question of multimodal alignment under known correspondence between modalities. Evaluation on experimentally measured diffraction data and unstructured literature text is an important next step rather than a claim of the present study.

\section{Results}
MatBind aligns four complementary material modalities --- crystal structure, \pxrd, \dos, and natural-language text --- into a single embedding space using contrastive learning.
In this section, we first outline the main design choices of the framework (illustrated in Figure~\ref{fig:overview}), then examine whether the learned representation captures physically meaningful structure, and finally evaluate its performance on direct cross-modal retrieval, emergent zero-shot retrieval, and multimodal retrieval in realistic characterization scenarios.

\subsection{Framework design}

Each modality requires a dedicated encoder whose inductive bias matches the structure of the underlying data. At the atomic scale, a crystalline material is naturally represented as a periodic graph in which atoms are nodes and interatomic interactions are edges; therefore, we employ a graph convolutional neural network (GCNN)~\cite{kipf2016semisupervised} as the encoder for crystal structures.
\pxrd\ patterns are one-dimensional intensity profiles over a fixed angular range and share characteristics with other sequential signals for which convolutional architectures have been proven effective; therefore, we adopt a ResNet-based encoder~\cite{he2015deep} 
Unlike \pxrd, \dos~ is defined over an energy axis whose range varies across materials; a fixed-size convolutional model would therefore be unsuitable. This motivates the choice of
a transformer~\cite{vaswani2017attention}-based model for \dos\ that treats
\dos\ as a variable-length sequence of (energy, density) pairs, with the energy
axis shifted relative to the Fermi level to enable comparison across materials.
Finally, textual descriptions of crystal structures are encoded using MatBERT~\cite{walker2021impact,Trewartha_2022}, a domain-specific BERT model pretrained on the materials science literature, which provides richer representations of compositional and symmetry information than a generic language model~\cite{Alampara2026}.

To align these encoders, we adopt the ImageBind~\cite{girdhar2023imagebind} framework, which uses contrastive learning~\cite{chen2020simple} to bring matched pairs closer while separating unmatched ones in the embedding space. In this setup, crystal structure serves as the central anchor modality: it is the most consistently available representation across public databases, and it encodes the physical information from which all other modalities ultimately derive. Pairwise contrastive objectives are applied between the crystal structure and each auxiliary modality; as we show below, this creates the alignment between auxiliary modalities
even when they are never explicitly paired during training. Full architectural
details and training settings are provided in Sections~\ref{sec:encoders}--\ref{sec:contrastive} and in Appendix~\ref{appdenix:model_para}.

\subsection{Representation of physical properties in the embedding space}

Before directly evaluating retrieval performance, we ask whether the shared embedding space organizes materials in a physically meaningful way. To address this question, we first analyze the text embeddings and annotate each point with its band gap and crystal system, properties that were \emph{not} included as training objectives.

\begin{figure}[h!]
    \centering
     \begin{subfigure}[b]{0.475\textwidth}
         \centering
         \includegraphics[width=\textwidth]{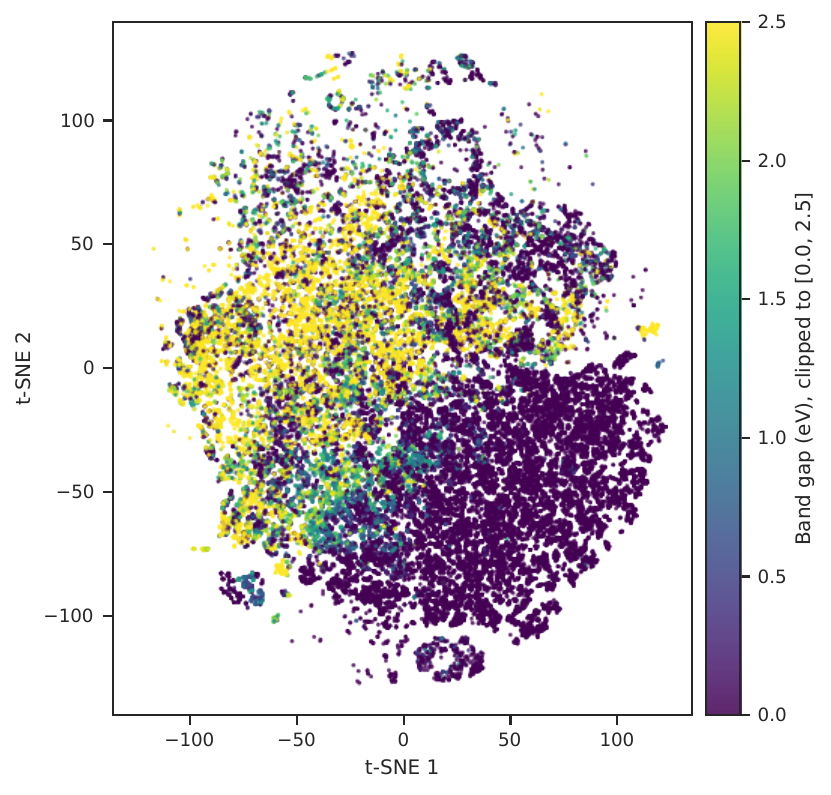}
         \caption{Colored by band gap}
         \label{fig:embedding_band_gap}
     \end{subfigure}
     \begin{subfigure}[b]{0.475\textwidth}
         \centering
         \includegraphics[width=\textwidth]{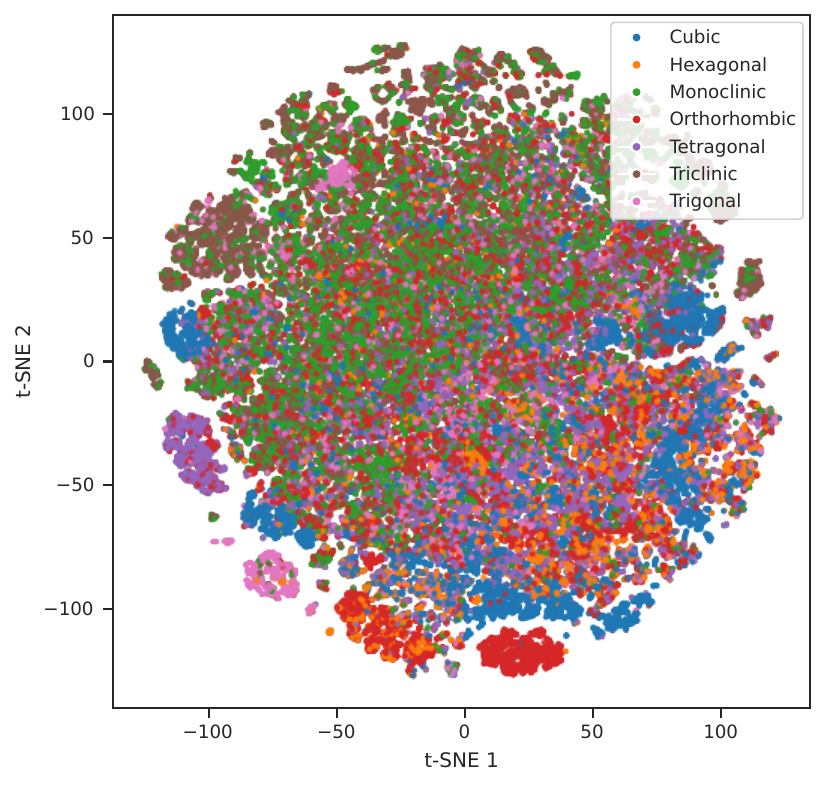}
         \caption{Colored by crystal system}
         \label{fig:embedding_crystal_system}
     \end{subfigure}
    \caption{\textbf{Text embeddings organize materials according to physically meaningful properties.}
    t-SNE visualization of the learned text embeddings.
    (a) Embeddings colored by band gap values (clipped at 2.5 for visualization), showing a smooth organization with respect to band gap.
    (b) Embeddings colored by crystal system, revealing weaker but still observable clustering according to crystallographic symmetry.
    }
    \label{fig:embedding_interpretation}
\end{figure}

The high-dimensional embeddings from text encoder are projected using t-SNE~\cite{van2008visualizing}, as shown in Figure~\ref{fig:embedding_interpretation}. UMAP~\cite{mcinnes2018umap} and PCA~\cite{pearson1901liii} projections, along with details of t-SNE, are provided in Appendix~\ref{UMAP_viusal}.
In Figure~\ref{fig:embedding_band_gap}, the embedding space is colored by band gap values (clipped at 2.5 for visualization), and materials with similar band gaps form distinct clusters. Figure~\ref{fig:embedding_crystal_system} presents the same embeddings labeled by crystal system, revealing weaker but still observable clustering behavior.

To further quantify this structure, we perform linear probing experiments. Linear probes can demonstrate whether label recovery is possible by drawing simple boundaries in the learned embedding space. We train a linear classifier on top of frozen modality embeddings to predict crystal system labels. A linear classifier is chosen for its limited capacity, since the encoder weights are frozen during probing. This ensures that probe accuracy reflects how much label-relevant structure is already linearly recoverable from the representation, rather than the capacity of the classifier itself. This task was not included in the training objective.

\begin{figure}[h]
    \centering
    \begin{subfigure}[t]{0.5\textwidth}
        \centering
        \includegraphics[width=\textwidth]{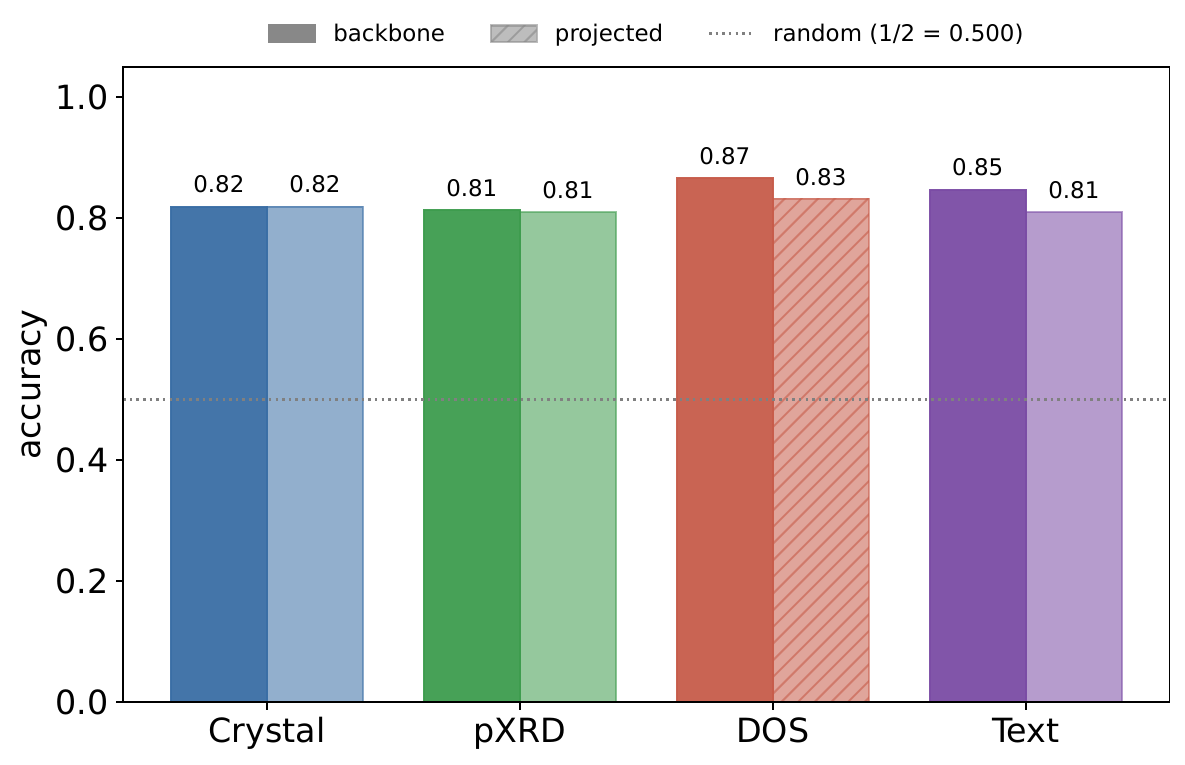}
        \caption{Metallicity classification across modalities and probe levels}
        \label{fig:linear_probe_metallicity}
    \end{subfigure}%
    \begin{subfigure}[t]{0.5\textwidth}
        \centering
        \includegraphics[width=\linewidth]{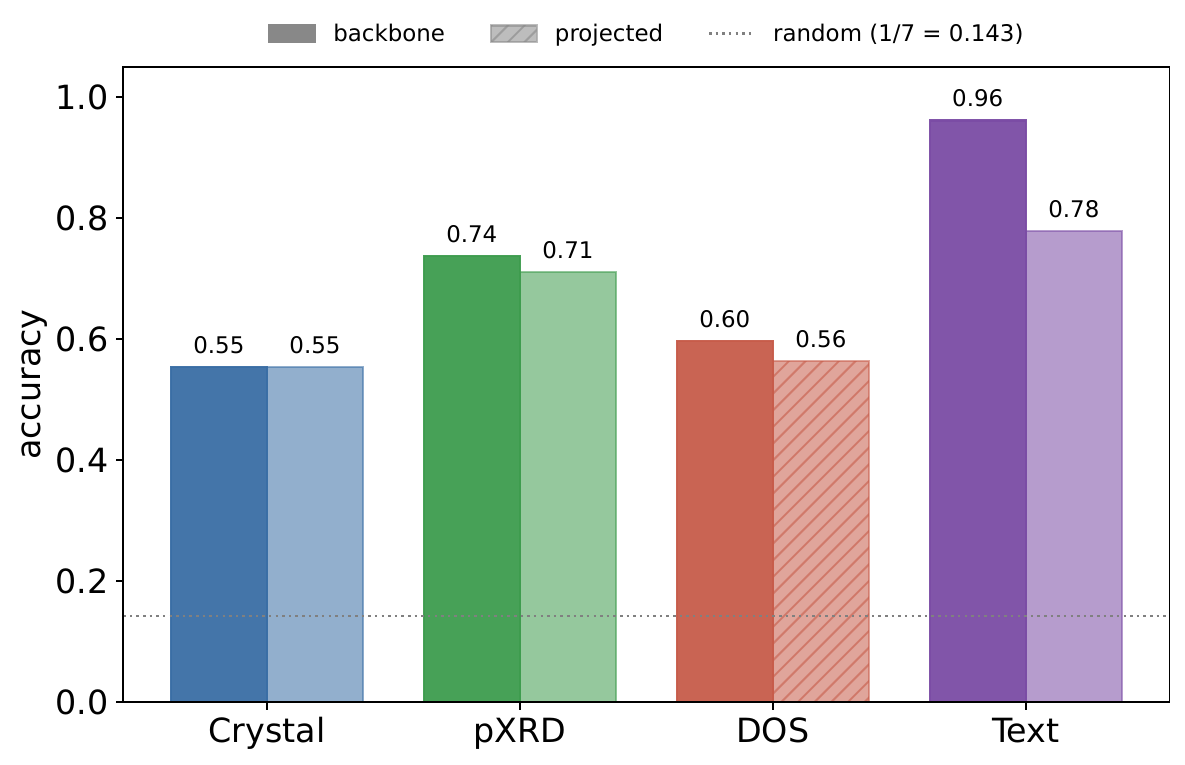}
        \caption{Crystal-system classification across modalities and probe levels}
        \label{fig:linear_probe_crystal_system}
    \end{subfigure}
    \caption{
    \textbf{MatBind embeddings encode material properties without explicit supervision.} The solid bar indicates the linear probe performed directly on the embedding (no projection head), and the hatched bar indicates the probe performed after the projection head.
    (a) \textbf{Metallicity classification across modalities and probe levels}. Backbone (solid) and projected (hatched) accuracy for each of the four MatBind encoders on the binary metallicity label.
    (b) \textbf{Crystal-system classification across modalities and probe levels}. Backbone (solid) and projected (hatched) accuracy for each of the four MatBind encoders on the 7-way crystal-system label. The dotted line at $1/7 \approx 0.143$ is a uniform-random chance
    }
    \label{fig:linear_probe}
\end{figure}

We prepare two tasks, each a linear probe experiment: a binary classification to identify metallicity and a multi-class classification to identify crystal system. Figure~\ref{fig:linear_probe_metallicity} shows the accuracy results for the metallicity classification. 

In crystal-system classification using the linear probe, which is shown in Figure~\ref{fig:linear_probe_crystal_system}, the text modality explicitly encodes crystal-system information; therefore, the text encoder achieves high accuracy. The information is propagated unevenly because the crystal structure and \dos~ modalities perform relatively poorly, even though they score well above the random-chance threshold of $0.143$.

In the Matbind architecture, the encoder backbone dimensionality is higher than the projected dimensionality, with a sole exception: the crystal structure modality, for which both backbone and projected dimensions are equal (128). And from Figure~\ref{fig:linear_probe}, the high-dimensional backbone performs better in nearly all cases, since it carries more information than the projected low-dimensional layer. This cannot be said for crystal modality since its backbone is too narrow to carry a symmetry signal at the same level as the wider partner encoders.

Having established that the resulting embedding space reflects physical properties, we now evaluate whether this structure supports accurate cross-modal retrieval.

\subsection{Cross-modal retrieval}

We quantitatively evaluate the learned embedding space using cross-modal retrieval tasks. Given a query in one modality, the model retrieves the closest match from a reference set of a second modality by comparing embeddings in the shared space. Performance is measured using Recall@k, which quantifies the fraction of queries for which the correct match appears within the top-$k$ retrieved candidates. Details of the evaluation protocol and training dataset are provided in Section~\ref{sec:dataset}

\begin{figure}[h]
    \centering
    \begin{subfigure}[t]{0.5\textwidth}
        \centering
        \includegraphics[width=\textwidth]{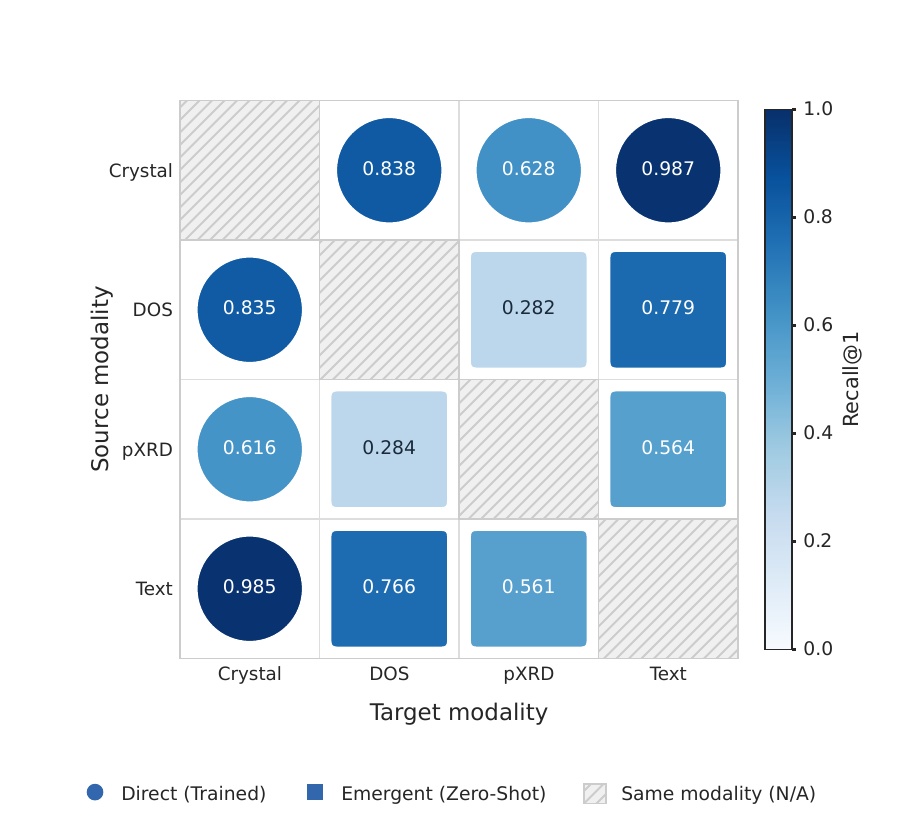}
        \caption{Recall@1 for different modality pairs}
        \label{fig:recall_at_1_matrix}
    \end{subfigure}%
    \begin{subfigure}[t]{0.5\textwidth}
        \centering
        \includegraphics[width=\linewidth]{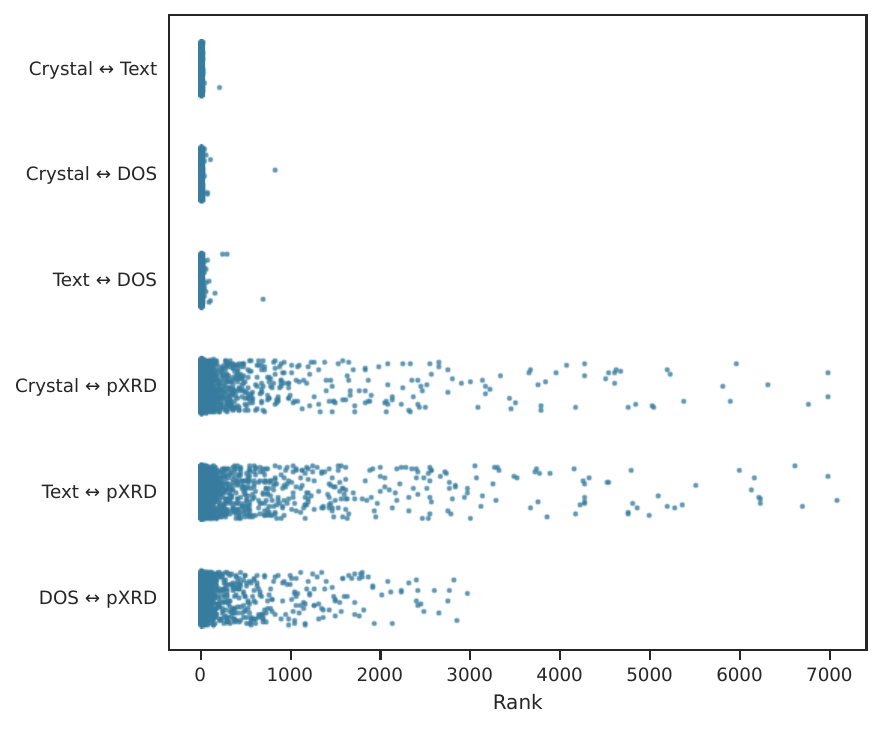}
        \caption{Distribution of retrieval ranks}
        \label{fig:strip_plot_all_rank}
    \end{subfigure}
    \caption{\textbf{Anchor-based alignment enables strong retrieval between both directly trained and emergent modality pairs.}
    (a) \textbf{Recall@1 for different modality pairs}. Circles denote modality pairs explicitly aligned during training, while rectangles indicate emergent (zero-shot) retrieval between modalities not directly paired. High performance in both cases demonstrates effective alignment, several emergent pairs achieve performance comparable to or exceeding directly trained pairs; notably, the emergent \dos$\leftrightarrow$text retrieval outperforms the directly trained crystal structure$\leftrightarrow$\pxrd.
    (b) \textbf{Distribution of retrieval ranks.} Lower ranks indicate better performance.
    Directly trained pairs involving crystal structure, \dos, and text retrieve
    the correct material at top ranks; emergent links follow closely.}
    \label{fig:retrieval_performance}
\end{figure}

Figure~\ref{fig:recall_at_1_matrix} shows Recall@1 results for all modality pairs and a model trained jointly on four modalities, using crystal structure as the central modality. Pairs that are explicitly aligned during training achieve consistently high retrieval performance, with crystal structure $\leftrightarrow$ text reaching near-perfect Recall@1 of $\approx$ 0.99, while crystal structure $\leftrightarrow$ DOS achieves strong Recall@1 of $\approx$ 0.84.

The crystal structure $\leftrightarrow$ \pxrd\ pair is a notable exception:
despite being a directly trained pair, its performance is much weaker relative to the others. This reflects the inherent many-to-one nature of \pxrd~:
multiple distinct crystal structures may result in nearly identical \pxrd~ patterns, which introduces ambiguity that the contrastive objective cannot fully resolve. Figure~\ref{fig:strip_plot_all_rank} further illustrates this situation: retrieval tasks involving \pxrd\ have ranks spread across a wide range, reflecting the inherent complexity of the task.

\subsection{Emergent zero-shot retrieval}
A key aspect of the ImageBind-style training scheme is that aligning all modalities to a common ``anchor'' modality induces alignment between auxiliary modalities that are \emph{never explicitly paired during training}. We refer to these as emergent retrieval links and evaluate the retrieval performance without going through the anchor modality. 

\begin{figure}[h]
    \centering
    \includegraphics[width=0.6\linewidth]{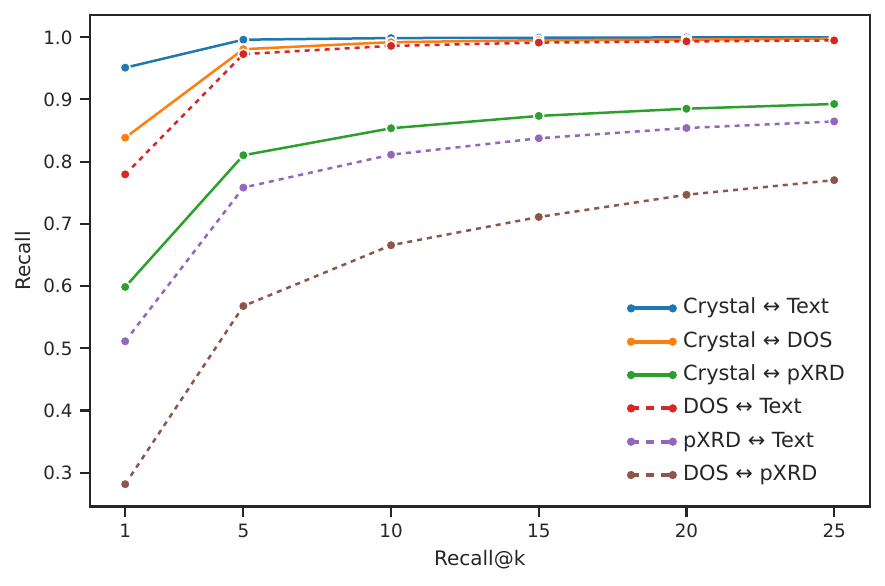}
    \caption{\textbf{Emergent retrieval can outperform directly trained modality pairs across the full recall curve.} Solid lines indicate directly trained pairs; dashed
    lines indicate emergent links. The emergent \dos$\leftrightarrow$text link
    surpasses the directly trained crystal structure$\leftrightarrow$\pxrd~
    pair across all $k$, demonstrating that the shared space captures
    cross-modal correspondence beyond explicit supervision.}
    \label{fig:performance_over_k}
\end{figure}

The main observation is that most emergent modality pairs achieve recall comparable to directly trained pairs. In particular, the emergent \dos$\leftrightarrow$text link performs substantially better than the directly trained crystal structure$\leftrightarrow$\pxrd\ pair across all values of $k$ (Figure~\ref{fig:performance_over_k}). This shows that the shared
embedding space consists of a genuine common representation rather than
merely memorizing pairwise correspondences. The shared space thus connects modalities through physics rather than through explicit pairing.

The only exception is the \dos-\pxrd~pair. The reduced performance in this case might cause by the fact that the limited shared information between \dos\ and \pxrd: the modalities represent fundamentally different physical aspects of materials (electronic structure versus long-range periodicity). The embedding space correctly reflects this physical reality.

\subsection{Practical application of multimodal retrieval}
\label{sec:cases}

In what follows, we demonstrate that the unified embedding space can be queried with  multiple modalities simultaneously, and that even partial auxiliary 
information is sufficient to resolve ambiguities that single-modality 
retrieval cannot.

\paragraph{Refined retrieval with complementary modalities}

Experimental characterization data, such as \pxrd~patterns, often lead to ambiguity in structure identification: the mapping from crystal structures to diffraction patterns is not one-to-one, and experimental measurements are further affected by noise, peak broadening, and incomplete sampling. As a consequence, retrieval based solely on \pxrd~frequently yields multiple candidate structures with similar similarity scores. 

The unified embedding space offers a natural remedy. In realistic characterization scenarios, auxiliary information is often available alongside a \pxrd~measurement, such as chemical composition, crystal system, or space group. These constraints can be incorporated directly as a complementary textual query. 
We evaluate this capability through a two-stage retrieval procedure: in the first stage, candidate structures are retrieved using only the \pxrd~pattern;  in the second stage, a partial textual description containing only composition, crystal system, and space group  is used to re-rank the candidates through a weighted scoring scheme:
\[
S = \alpha S_{\mathrm{\pxrd}} + (1 - \alpha) S_{\mathrm{text}},
\]
where $S_{\mathrm{\pxrd}}$ and $S_{\mathrm{text}}$ denote similarity scores from \pxrd-based and text-based retrieval, respectively, and $\alpha$ controls their relative contributions. 

\begin{figure}[h]
\centering
\includegraphics[width = \textwidth]{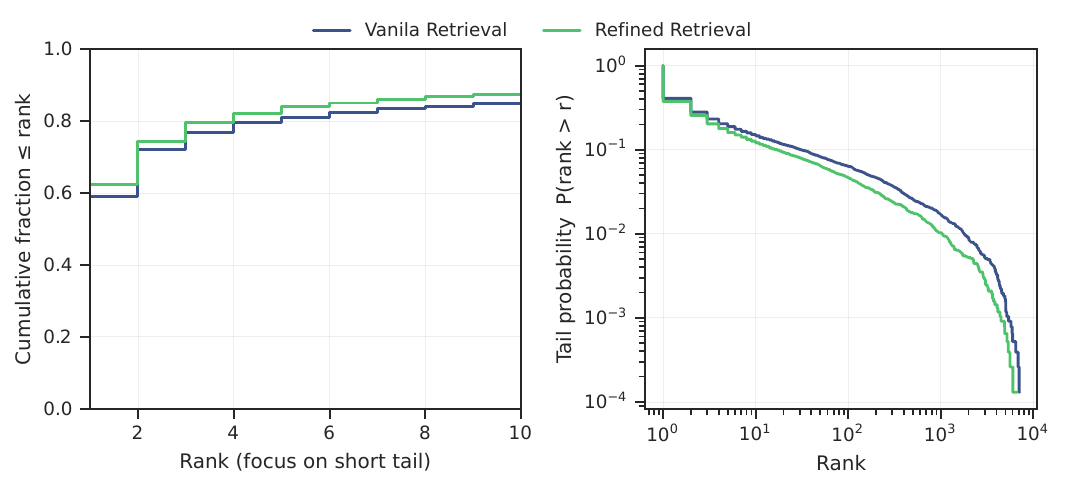}
\caption{\textbf{Combining complementary modalities consistently improves retrieval accuracy.}
Cumulative distribution (left) and complementary cumulative distribution (right) of retrieval ranks for retrieval without complementary information and retrieval refined by complementary information. The refined method concentrates retrievals at lower ranks and exhibits a faster-decaying tail, indicating improved overall retrieval performance.}
\label{fig:refined_retrieval}
\end{figure}

Despite the limited textual input, the refined retrieval consistently outperforms \pxrd-only retrieval: Recall@1 improves from $0.59$ to $0.62$ and Recall@5 from $0.81$ to $0.84$, demonstrating that even minimal auxiliary information can enhance retrieval accuracy.
Figure~\ref{fig:refined_retrieval} further shows that the rank distribution of the correct structure shifts toward lower ranks after refinement, with a fast-decaying tail indicating fewer catastrophic failures. From a physical perspective, this improvement reflects the complementary character of the two modalities: \pxrd\ encodes long-range periodic structure, while textual descriptors provide explicit constraints on composition and symmetry, which together reduce the ambiguity among candidate structures in the
embedding space.

\begin{figure*}[h]
    \centering
    \includegraphics[width = \linewidth]{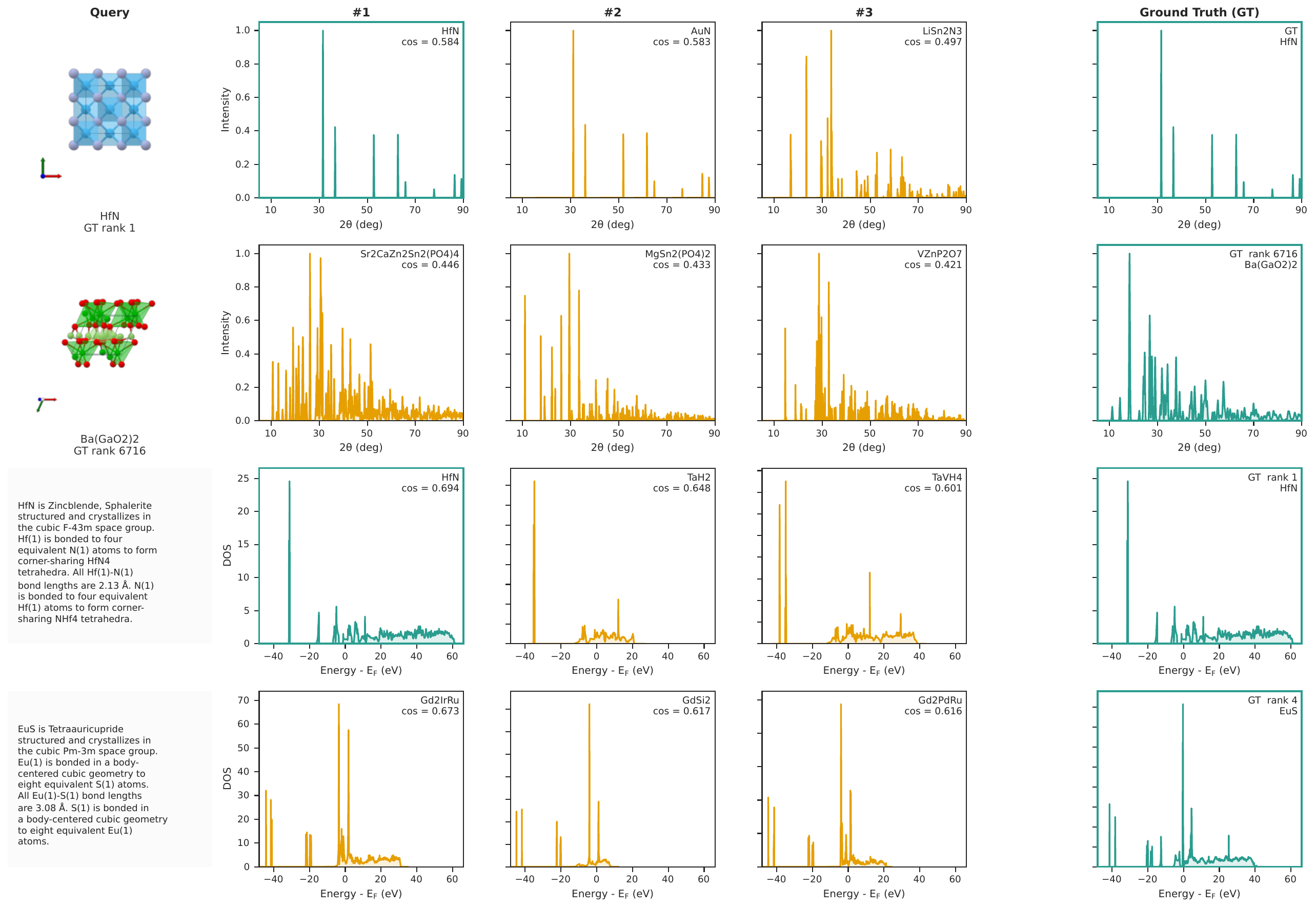}
    \caption{\textbf{The shared embedding space retrieves physically corresponding materials across modalities.} Representative retrievals from the shared embedding space, spanning successful and challenging cases. Green indicates the ground truth, and yellow indicates the top-3 retrieved results; GT rank refers to the position of the ground truth within the full retrieval ranking. The first two rows show retrieval from crystal structure to \pxrd, and the last two rows show retrieval from text to \dos. Cosine similarity scores are annotated in each subfigure and indicate similarity in the embedding space.}
    \label{fig:qualitative}
\end{figure*}

Figure~\ref{fig:qualitative} illustrates representative retrieval examples from the learned embedding space. These examples demonstrate that the framework recovers physically meaningful matches across heterogeneous modalities, including emergent pairs that were never explicitly aligned during training.

\section{Discussion}

\paragraph{Physical interpretation of retrieval performance}
The performance differences observed across modality pairs reflect underlying physical relationships. Crystal structure, \dos, and text are tightly coupled: the electronic structure of a material is determined by its atomic arrangement, and textual descriptions explicitly encode compositional and symmetry information derived from the same source. \pxrd, by contrast, is an inherently ambiguous modality --- the mapping from structure to diffraction pattern is many-to-one, meaning that distinct structures can produce near-identical patterns. The embedding space correctly reflects this physical reality: retrieval performance degrades precisely where physics predicts ambiguity, and improves when complementary information is added. This is not a limitation of the framework but a validation of it.

The ablation study further shows that this broad multimodal alignment comes with a measurable trade-off: optimizing a single shared space across all modalities can reduce the performance of specific modality pairs, most visibly crystal structure $\leftrightarrow$ pXRD, compared with a model trained only on that pair. This is expected because the embedding space must balance pairwise discrimination against global compatibility across modalities. The all-modality model should therefore not be interpreted as the optimal specialist model for every pairwise task, but as a general representation that enables interoperability, emergent zero-shot links, and multimodal querying.

\paragraph{Alignment versus modality-specific information}
The linear probing experiments reveal a trade-off between cross-modal alignment and the preservation of modality-specific information: fine-tuning the text encoder to align with other modalities partially overwrites the symmetry-related information encoded by MatBERT's domain-specific pretraining. This suggests that the contrastive objective, while effective
for retrieval, may compress modality-specific features that are useful for downstream tasks. Future work could explore whether this trade-off can be mitigated through, for example, auxiliary reconstruction objectives or modality-specific regularization terms that preserve encoder expressivity while maintaining cross-modal alignment.

\paragraph{Limitations}
The current framework has several limitations worth acknowledging. The dataset is sourced entirely from the Materials Project, which introduces a bias toward computationally stable, well-characterized materials; performance on experimentally synthesized materials with partial or noisy characterization data remains to be evaluated. Furthermore, while the framework is designed to handle missing modalities gracefully, the quality of emergent retrieval links depends on the informativeness of the anchor modality and the size of the training set. Because contrastive learning maximizes mutual information across paired modalities, modality-specific information may also be lost during alignment.

\paragraph{Extensibility to new modalities}
A natural question is how the framework generalizes to modalities not seen during training, for example, Raman spectra, optical absorption, or scanning tunneling microscopy images. In principle, the ImageBind-style architecture supports the addition of new modalities by training a new encoder against the existing crystal structure anchor, without retraining the full model. Performance after adding a new modality depends on both the alignment strategy and the characteristics of the modality. If the existing backbone is frozen, training should primarily bring the new modality embedding space closer to the existing shared space, with the binding strength depending on how strongly the new modality is related to the anchor modality. Ambiguous modalities, such as elasticity tensor, may therefore be more difficult to align effectively.

\paragraph{Towards a unified representation of materials knowledge}
A recurring challenge in materials science is that experimental, computational, and textual data are collected and stored in isolation, even though they all describe the same physical reality. This fragmentation reflects a mismatch between how materials data is organized and how physical understanding actually works. Different modalities do not contain different ``truths''; they reveal different facets of the same underlying system. A diffraction pattern, an electronic structure, and a textual description are complementary projections of a single physical object, yet standard practice treats them as separate data sources and, in doing so, discards information that is always present, even if not always measured.

This is precisely what materials scientists do intuitively: when characterizing an unknown material, one does not treat a diffraction measurement in isolation but immediately contextualizes it with knowledge of the likely composition, symmetry class, or electronic behavior. MatBind formalizes this intuition computationally, and the shared embedding space can be understood as a learned prior over the joint distribution of materials representations.

As the volume and diversity of materials data grow, multimodal representations will become increasingly essential for connecting experimental observations, computational predictions, and domain knowledge into a coherent whole.
 
\section{Conclusion}

We introduced MatBind, a contrastive learning framework that aligns four complementary materials modalities --- crystal structure, \pxrd, \dos, and text --- into a unified embedding space. Using crystal structure as a physical anchor, the framework induces alignment between modalities that are never explicitly paired during training, enabling zero-shot cross-modal retrieval as an emergent property of the shared representation. Retrieval performance improves systematically when modalities are combined, and the embedding space organizes materials according to physically meaningful properties without explicit supervision. These results demonstrate that treating heterogeneous materials data as complementary facets of a single physical system, rather than as isolated data sources, is a physically well-motivated modeling choice.
\section{Methods}

\subsection{Dataset}
\label{sec:dataset}
We curate our dataset from the Materials Project database~\cite{jain2013commentary}, which comprises roughly \num{155361} unique materials. For each entry, we retrieve the crystal structure and the corresponding \dos~and we restrict the \dos~to the total density of states. 

Textual descriptions are produced using 
Robocrystallographer~\cite{ganose2019robocrystallographer}. \pxrd~patterns are simulated using \verb|pymatgen|~\cite{pymatgen}, with diffraction peaks  evaluated over a $2\theta$ interval of $[\SI{5}{\degree}, \SI{90}{\degree}]$. The ideal peak positions are broadened with Gaussian profiles, with the full 
width at half maximum determined via the Scherrer equation~\cite{scherrer_eq}. To introduce realistic variability, the average crystallite size is randomly drawn from $[\SI{200}{\mathring{\text{A}}}, \SI{1000}{\mathring{\text{A}}}]$. The resulting \pxrd~intensity profiles are sampled with a step size of $\SI{0.01}{\degree}$, 
yielding vectors of length \num{8501}, and the intensities are scaled to $[0, 1]$.

Not every material in the database is associated with all modalities; every entry has a crystal structure and a simulated \pxrd~pattern, while \dos~and text are available for subsets only. The dataset composition and train/validation splits are summarized in Table~\ref{tab:dataset_stats}. Because we do not perform hyperparameter optimization, we do not create a separate validation/test split; instead, we use validation as a test set that is never exposed to the model during training.

\begin{table}[h]
    \centering
    \caption{\textbf{Dataset statistics.} Sample counts per modality and 
    data split. Crystal structure and \pxrd~are available for all samples; 
    \dos~and text are available for subsets only.}
    \label{tab:dataset_stats}
    \begin{tabular}{lcccc}
        \hline
        Split & Crystal / \pxrd & \dos & Text \\
        \hline
        Train      & \num{147592} & \num{59847} & 
                     \num{146990} \\
        Validation & \num{7769}                 & \num{3193}                 & 
                     \num{7659}                 \\
        \hline
        Total      & \num{155361}               & \num{63040}                & 
                     \num{154649}               \\
        \hline
    \end{tabular}
\end{table}

\subsection{Encoders}
\label{sec:encoders}

\paragraph{Crystal structure encoder}
A crystal structure is naturally represented as a periodic graph, with atoms as nodes and interatomic interactions as edges. We
therefore employ a graph convolutional neural network (GCNN)~\cite{kipf2016semisupervised} with six stacked convolutional
layers as the crystal structure encoder. Node features are constructed using a site-specific atomic encoding scheme~\cite{taniai2024crystalformer}, where single-species sites are represented by binary vectors and mixed-occupancy sites by weighted combinations. Edges are defined based on nearest neighbors within a cutoff radius, and interatomic distances are encoded using a Gaussian radial basis expansion with 41 components.

\paragraph{\pxrd\ encoder}

The \pxrd~ encoder is based on a convolutional ResNet architecture, which has previously demonstrated strong performance on diffraction-based tasks such as space group prediction~\cite{Schopmans_2023}. The model can be pretrained on synthetic \pxrd~data generated from crystal structures, optionally with added noise to mimic experimental conditions. In this work, we adopt this architecture to encode \pxrd~patterns into a latent representation compatible with other modalities.

\paragraph{\dos\ encoder}

Unlike \pxrd, which is sampled over a fixed angular range, the \dos\ is defined over an energy axis whose range varies between materials, making fixed-size convolutional architectures unsuitable. We therefore employ a Transformer-based encoder~\cite{vaswani2017attention} that treats \dos\ as a variable-length sequence of (energy, density) pairs. Following \cite{moro2025multimodal}, the energy axis is shifted by subtracting the Fermi energy, enabling the model to learn relative energy-dependent features that are comparable across materials. Positional embeddings and explicit energy values are both included as input features, allowing the Transformer to capture both local spectral features and global electronic structure.

\paragraph{Text encoder}

Textual descriptions are tokenized using the BERT tokenizer and encoded using MatBERT, a domain-specific BERT model pretrained on scientific literature~\cite{devlin2019bert}. The final text embedding is obtained by computing an average of the token embeddings from the last hidden layer, excluding padding tokens. Additionally, because the pretrained MatBERT model has a fixed context window of 512 tokens, tokenized text is truncated to 512 tokens, which affects around 40\% of the text data, particularly materials with many elements.

\subsection{Contrastive alignment}
\label{sec:contrastive}

MatBind follows the ImageBind framework~\cite{girdhar2023imagebind} to align multiple modalities within a shared embedding space. Crystal structure ($\mathcal{C}$) is used as the anchor modality and is paired with each auxiliary modality ($\mathcal{E} \in \{\text{text}, \text{\dos}, \text{\pxrd}\}$).

Each encoder maps its input to a latent representation:
\[
\mathbf{a}_i = \phi_{\mathcal{C}}(a_i), \quad \mathbf{b}_i = \phi_{\mathcal{E}}(b_i),
\]
where $(a_i, b_i)$ denotes a matched pair.

We optimize a contrastive objective that brings matching pairs closer while pushing non-matching pairs apart. This is implemented using the InfoNCE loss~\cite{oord2018representation}:

\begin{equation}
   L = -\frac{1}{n} \sum_i 
   \log 
   \frac{\exp(\mathbf{a}_i^\top \mathbf{b}_i / \tau)}
   {\sum_j \exp(\mathbf{a}_i^\top \mathbf{b}_j / \tau)}.
\end{equation}

Here, $\tau$ is a temperature parameter controlling the sharpness of similarity scores. For our training, it is fixed at 0.07, following CLIP. Negative examples are implicitly provided by other samples within the batch.

\section*{Acknowledgements}

This work has been supported by the Helmholtz Association within the framework of the Helmholtz Foundation Model Initiative (project SOL-AI). It was also supported by the Helmholtz AI computing resources (HAICORE) of the Helmholtz Association's Initiative and Networking Fund on the HAICORE@FZJ partition.
K.M.J.\ is a member of the NFDI consortium FAIRmat - Deutsche Forschungsgemeinschaft (DFG) - Project 460197019. 

The authors gratefully acknowledge the Gauss Centre for Supercomputing e.V. (www.gauss-centre.eu) for funding this project by providing computing time through the John von Neumann Institute for Computing (NIC) on the GCS Supercomputer JUWELS at Jülich Supercomputing Centre (JSC). Additionally, computational resources were provided by the German AI Service Center WestAI.

\section*{Author contributions}
\footnotesize
\insertcredits
\normalsize 

\section*{Conflicts of interests}
The authors declare no conflicts of interest.

\section*{Data and code availability}
Code is available on GitHub (\url{https://github.com/HFMI-SOL-AI/MatBind}).

The datasets generated and analysed during the current study are available in the HuggingFace repository (\url{https://huggingface.co/datasets/SOL-AI/matbind-data})

\section*{Declaration of Generative AI and AI-assisted technologies in the writing process}
During the preparation of this work, the authors used OpenAI's GPT, Anthropic’s Claude and Google's Gemini models to improve language and readability. After using this service, the authors reviewed and edited the content as needed and take full responsibility for the content of the publication.

\printbibliography

@article{koker2022graph,
  author       = {Koker, Teddy and Quigley, Keegan and Spaeth, Will and Frey, Nathan C and Li, Lin},
  date         = {2022},
  journaltitle = {arXiv preprint arXiv:2211.13408},
  title        = {Graph contrastive learning for materials},
}

@article{xu2025kan,
  author       = {Xu, Chenlei and Su, Tianhao and Xiong, Jie and Wu, Yue and Dong, Shuya and Jiang, Tian and He,
                  Mengwei and Chen, Shuai and Zhang, Tong-Yi},
  date         = {2025},
  journaltitle = {arXiv preprint arXiv:2511.04055},
  title        = {KAN-Enhanced Contrastive Learning Accelerating Crystal Structure Identification from XRD Patterns},
}

@inproceedings{chen2020simple,
  author       = {Chen, Ting and Kornblith, Simon and Norouzi, Mohammad and Hinton, Geoffrey},
  organization = {PmLR},
  booktitle    = {International conference on machine learning},
  date         = {2020},
  pages        = {1597--1607},
  title        = {A simple framework for contrastive learning of visual representations},
}

@inproceedings{radford2021learning,
  author       = {Radford, Alec and Kim, Jong Wook and Hallacy, Chris and Ramesh, Aditya and Goh, Gabriel and Agarwal,
                  Sandhini and Sastry, Girish and Askell, Amanda and Mishkin, Pamela and Clark, Jack and others},
  organization = {PmLR},
  booktitle    = {International conference on machine learning},
  date         = {2021},
  pages        = {8748--8763},
  title        = {Learning transferable visual models from natural language supervision},
}

@article{park2025contrastive,
  author       = {Park, Yang Jeong and Kumaran, Mayank and Hsu, Chia-Wei and Olivetti, Elsa and Li, Ju},
  date         = {2025},
  journaltitle = {arXiv preprint arXiv:2502.16451},
  title        = {Contrastive learning of English language and crystal graphs for multimodal representation of
                  materials knowledge},
}

@article{suzuki2025bridging,
  author       = {Suzuki, Yuta and Taniai, Tatsunori and Igarashi, Ryo and Saito, Kotaro and Chiba, Naoya and Ushiku,
                  Yoshitaka and Ono, Kanta},
  publisher    = {IOP Publishing},
  date         = {2025},
  journaltitle = {Machine Learning: Science and Technology},
  number       = {3},
  pages        = {035006},
  title        = {Bridging text and crystal structures: literature-driven contrastive learning for materials science},
  volume       = {6},
}

@article{kipf2016semisupervised,
  author       = {Kipf, Thomas and Welling, M.},
  date         = {2016},
  journaltitle = {International Conference on Learning Representations},
  title        = {Semi-Supervised Classification with Graph Convolutional Networks},
}

@inproceedings{taniai2024crystalformer,
  author    = {Taniai, Tatsunori and Igarashi, Ryo and Suzuki, Yuta and Chiba, Naoya and Saito, Kotaro and Ushiku,
               Yoshitaka and Ono, Kanta},
  url       = {https://openreview.net/forum?id=fxQiecl9HB},
  booktitle = {The Twelfth International Conference on Learning Representations},
  date      = {2024},
  title     = {Crystalformer: Infinitely Connected Attention for Periodic Structure Encoding},
}

@article{he2015deep,
  author       = {He, Kaiming and Zhang, X. and Ren, Shaoqing and Sun, Jian},
  date         = {2015},
  doi          = {10.1109/cvpr.2016.90},
  journaltitle = {Computer Vision and Pattern Recognition},
  title        = {Deep Residual Learning for Image Recognition},
}

@article{Schopmans_2023,
  author       = {Schopmans, Henrik and Reiser, Patrick and Friederich, Pascal},
  publisher    = {Royal Society of Chemistry (RSC)},
  url          = {http://dx.doi.org/10.1039/D3DD00071K},
  date         = {2023},
  doi          = {10.1039/d3dd00071k},
  issn         = {2635-098X},
  journaltitle = {Digital Discovery},
  number       = {5},
  pages        = {1414--1424},
  title        = {Neural networks trained on synthetically generated crystals can extract structural information from
                  ICSD powder X-ray diffractograms},
  volume       = {2},
}

@article{vaswani2017attention,
  author       = {Vaswani, A},
  date         = {2017},
  journaltitle = {Advances in Neural Information Processing Systems},
  title        = {Attention is all you need},
}

@article{moro2025multimodal,
  author       = {Moro, Viggo and Loh, Charlotte and Dangovski, Rumen and Ghorashi, Ali and Ma, Andrew and Chen, Zhuo
                  and Kim, Samuel and Lu, Peter Y and Christensen, Thomas and Soljačić, Marin},
  publisher    = {Elsevier},
  date         = {2025},
  journaltitle = {Newton},
  number       = {1},
  title        = {Multimodal foundation models for material property prediction and discovery},
  volume       = {1},
}

@article{walker2021impact,
  author       = {Walker, Nicholas and Trewartha, Amalie and Huo, Haoyan and Lee, Sanghoon and Cruse, Kevin and
                  Dagdelen, John and Dunn, Alexander and Persson, Kristin and Ceder, Gerbrand and Jain, Anubhav},
  date         = {2021},
  journaltitle = {Available at SSRN 3950755},
  title        = {The Impact of Domain-Specific Pre-Training on Named Entity Recognition Tasks in Materials Science},
}

@article{Trewartha_2022,
  author       = {Trewartha, Amalie and Walker, Nicholas and Huo, Haoyan and Lee, Sanghoon and Cruse, Kevin and
                  Dagdelen, John and Dunn, Alexander and Persson, Kristin A. and Ceder, Gerbrand and Jain, Anubhav},
  publisher    = {Elsevier BV},
  url          = {http://dx.doi.org/10.1016/j.patter.2022.100488},
  date         = {2022-04},
  doi          = {10.1016/j.patter.2022.100488},
  issn         = {2666-3899},
  journaltitle = {Patterns},
  number       = {4},
  pages        = {100488},
  title        = {Quantifying the advantage of domain-specific pre-training on named entity recognition tasks in
                  materials science},
  volume       = {3},
}

@article{devlin2019bert,
  author       = {Devlin, Jacob and Chang, Ming-Wei and Lee, Kenton and Toutanova, Kristina},
  date         = {2019},
  doi          = {10.18653/v1/N19-1423},
  journaltitle = {North American Chapter of the Association for Computational Linguistics},
  title        = {BERT: Pre-training of Deep Bidirectional Transformers for Language Understanding},
}

@article{girdhar2023imagebind,
  author       = {Girdhar, Rohit and El-Nouby, Alaaeldin and Liu, Zhuang and Singh, Mannat and Alwala, Kalyan Vasudev
                  and Joulin, Armand and Misra, Ishan},
  date         = {2023},
  journaltitle = {arXiv preprint arXiv: 2305.05665},
  title        = {ImageBind: One Embedding Space To Bind Them All},
}

@article{oord2018representation,
  author       = {van den Oord, Aaron and Li, Yazhe and Vinyals, Oriol},
  date         = {2018},
  journaltitle = {arXiv preprint arXiv:1807.03748},
  title        = {Representation Learning with Contrastive Predictive Coding},
}

@article{jain2013commentary,
  author       = {Jain, Anubhav and Ong, Shyue Ping and Hautier, Geoffroy and Chen, Wei and Richards, William Davidson
                  and Dacek, Stephen and Cholia, Shreyas and Gunter, Dan and Skinner, David and Ceder, Gerbrand and
                  others},
  publisher    = {AIP Publishing},
  date         = {2013},
  journaltitle = {APL materials},
  number       = {1},
  title        = {Commentary: The Materials Project: A materials genome approach to accelerating materials innovation},
  volume       = {1},
}

@article{ganose2019robocrystallographer,
  author       = {Ganose, Alex M and Jain, Anubhav},
  publisher    = {Cambridge University Press},
  date         = {2019},
  journaltitle = {MRS Communications},
  number       = {3},
  pages        = {874--881},
  title        = {Robocrystallographer: automated crystal structure text descriptions and analysis},
  volume       = {9},
}

@article{pymatgen,
  author       = {Ong, Shyue Ping and Richards, William Davidson and Jain, Anubhav and Hautier, Geoffroy and Kocher,
                  Michael and Cholia, Shreyas and Gunter, Dan and Chevrier, Vincent L. and Persson, Kristin A. and
                  Ceder, Gerbrand},
  url          = {https://www.sciencedirect.com/science/article/pii/S0927025612006295},
  date         = {2013},
  doi          = {https://doi.org/10.1016/j.commatsci.2012.10.028},
  issn         = {0927-0256},
  journaltitle = {Computational Materials Science},
  pages        = {314--319},
  title        = {Python Materials Genomics (pymatgen): A robust, open-source python library for materials analysis},
  volume       = {68},
}

@article{scherrer_eq,
  author       = {Patterson, A. L.},
  publisher    = {American Physical Society},
  url          = {https://link.aps.org/doi/10.1103/PhysRev.56.978},
  date         = {1939-11},
  doi          = {10.1103/PhysRev.56.978},
  issue        = {10},
  journaltitle = {Phys. Rev.},
  pages        = {978--982},
  title        = {The Scherrer Formula for X-Ray Particle Size Determination},
  volume       = {56},
}

@article{van2008visualizing,
  author       = {Van der Maaten, Laurens and Hinton, Geoffrey},
  date         = {2008},
  journaltitle = {Journal of machine learning research},
  number       = {11},
  title        = {Visualizing data using t-SNE.},
  volume       = {9},
}

@article{mcinnes2018umap,
  author       = {McInnes, Leland and Healy, John and Melville, James},
  date         = {2018},
  journaltitle = {arXiv preprint arXiv:1802.03426},
  title        = {Umap: Uniform manifold approximation and projection for dimension reduction},
}

@article{pearson1901liii,
  title={LIII. On lines and planes of closest fit to systems of points in space},
  author={Pearson, Karl},
  journal={The London, Edinburgh, and Dublin philosophical magazine and journal of science},
  volume={2},
  number={11},
  pages={559--572},
  year={1901},
  publisher={Taylor \& Francis}
}

@article{Alampara2026,
  title = {General-Purpose Models for the Chemical Sciences: LLMs and Beyond},
  volume = {126},
  ISSN = {1520-6890},
  url = {http://dx.doi.org/10.1021/acs.chemrev.5c00583},
  DOI = {10.1021/acs.chemrev.5c00583},
  number = {4},
  journal = {Chemical Reviews},
  publisher = {American Chemical Society (ACS)},
  author = {Alampara,  Nawaf and Aneesh,  Anagha and Ríos-García,  Martiño and Mirza,  Adrian and Schilling-Wilhelmi,  Mara and Aghajani,  Ali Asghar and Sun,  Meiling and Prastalo,  Gordan and Jablonka,  Kevin Maik},
  year = {2026},
  month = Feb,
  pages = {2484–2549}
}

\appendix
\section{Related Work}
\label{sec:related_work}
\paragraph{Contrastive learning}
Since the introduction of SimCLR \cite{chen2020simple}, contrastive learning has gained significant traction in unsupervised representation learning due to its ability to produce highly structured and transferable latent spaces. Building on its success in unimodal settings, the paradigm has been extended to multimodal scenarios, where positive and negative pairs are constructed across different modalities rather than within the same one, such as images, text, video, and audio. This cross-modal alignment requires modifications to the standard contrastive setup, as demonstrated by CLIP \cite{radford2021learning}, which learns joint representations by aligning images and text in a shared embedding space. Further extending this idea, ImageBind \cite{girdhar2023imagebind} generalizes the framework to a unified multimodal setting that jointly models seven distinct modalities, highlighting the scalability and flexibility of contrastive objectives for learning aligned representations across heterogeneous data sources.
\paragraph{Contrastive learning in material science}
Given the strong performance of contrastive learning in unsupervised representation learning, materials science stands to benefit substantially from this paradigm. Recent efforts have begun to adapt SimCLR-like frameworks to domain-specific data, for example, by applying augmentations to crystal structures and learning structure-aware embeddings through instance discrimination \cite{koker2022graph}. Beyond unimodal structure-based learning, incorporating additional modalities has emerged as a promising direction. For instance, some studies aim to align crystal structures with powder X-ray diffraction (\pxrd) patterns to bridge structural and experimental representations \cite{xu2025kan}. In parallel, inspired by advances in multimodal models such as CLIP, recent studies explore aligning crystal structures with textual descriptions by leveraging large language models, thereby connecting atomistic information with semantic knowledge \cite{park2025contrastive,suzuki2025bridging}. These developments suggest that contrastive learning can serve as a unifying framework for integrating heterogeneous materials data into a shared latent space.

\section{Additional visualization of embedding space}
\label{UMAP_viusal}
For dimensionality reduction, we use all materials in the dataset as inputs and subsequently annotate them with the available properties. Consequently, the number of points in the band gap plot is smaller than in the crystal system plot because band gap data are not always available, as noted in Section~\ref{sec:dataset}.
\begin{figure}[!h]
    \centering
     \begin{subfigure}[b]{0.45\textwidth}
         \centering
         \includegraphics[width=\textwidth]{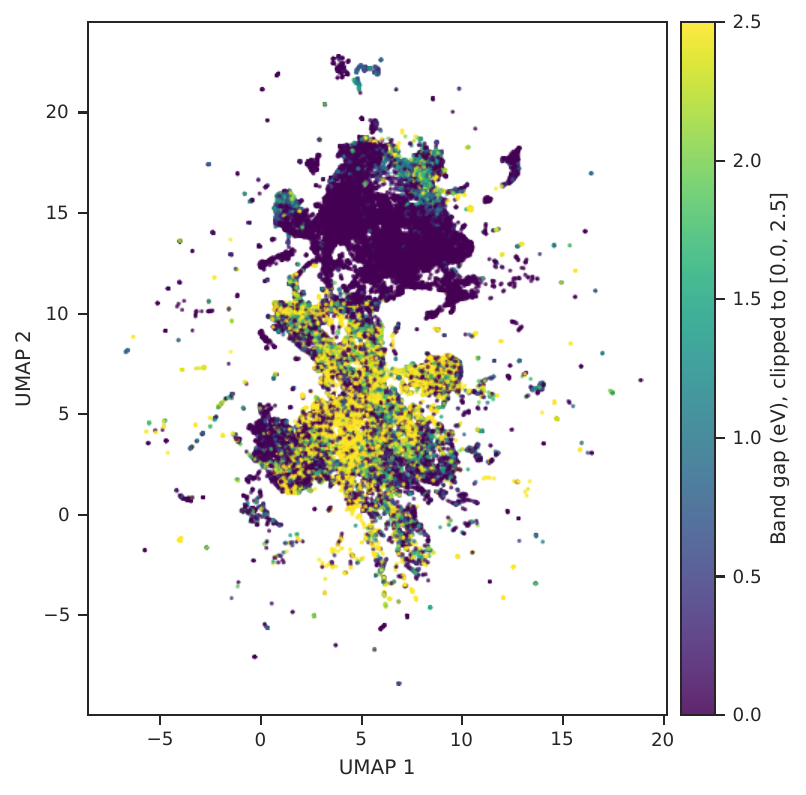}
         \caption{Colored by band gap}
     \end{subfigure}
     \begin{subfigure}[b]{0.45\textwidth}
         \centering
         \includegraphics[width=\textwidth]{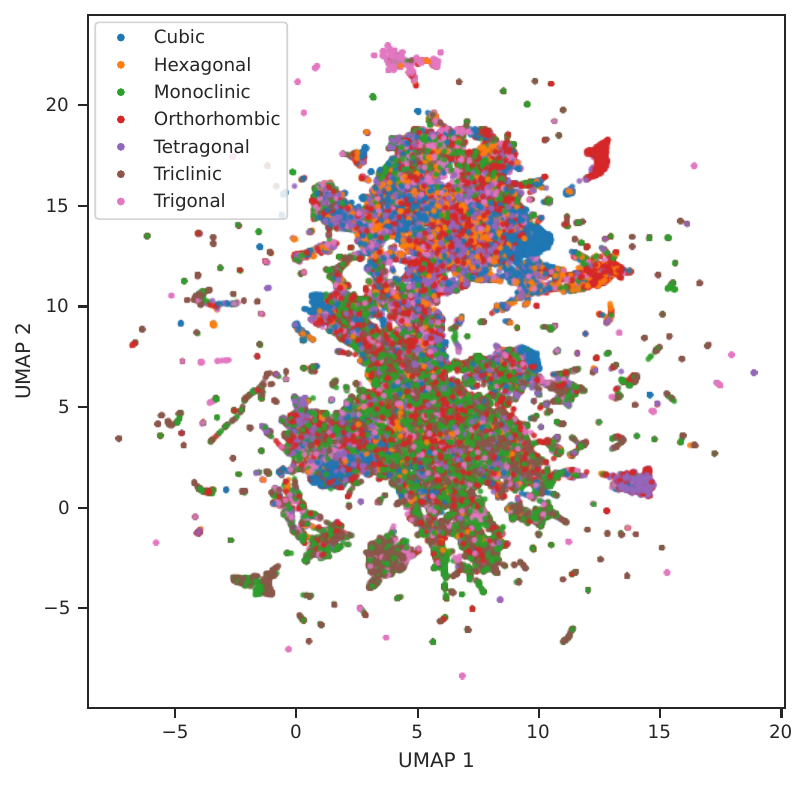}
         \caption{Colored by crystal system}
     \end{subfigure}
    \caption{\textbf{Interpretability of text embeddings visualized using UMAP.} The number of neighbors is set to 15, the minimum distance is 0.1, and the metric is cosine.}
    \label{fig:embedding_umap}
\end{figure}

\begin{figure}[!h]
    \centering
     \begin{subfigure}[b]{0.45\textwidth}
         \centering
         \includegraphics[width=\textwidth]{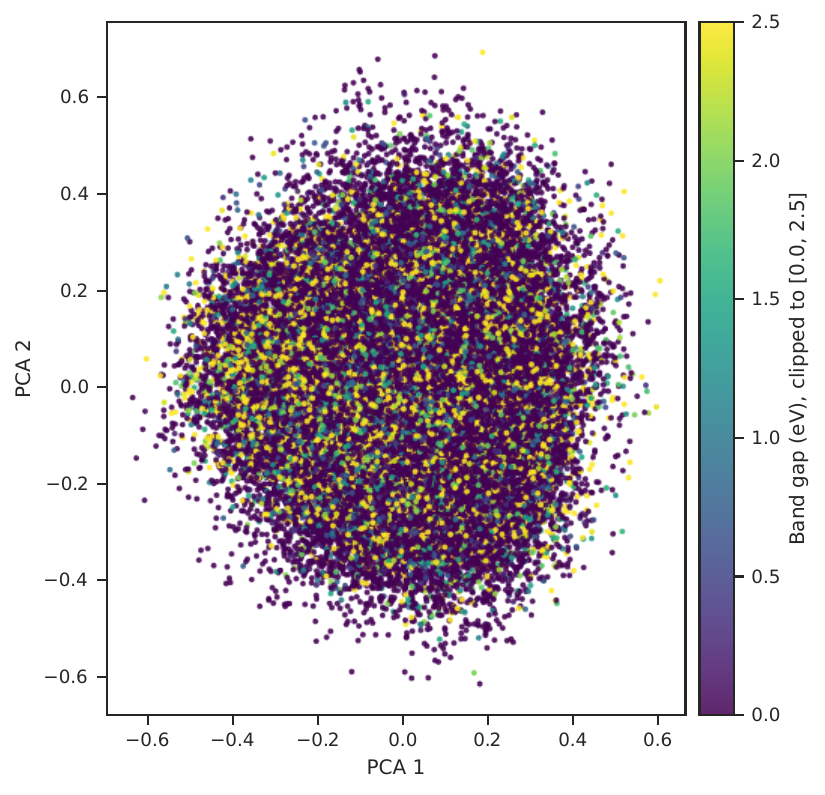}
         \caption{Colored by band gap}
     \end{subfigure}
     \begin{subfigure}[b]{0.45\textwidth}
         \centering
         \includegraphics[width=\textwidth]{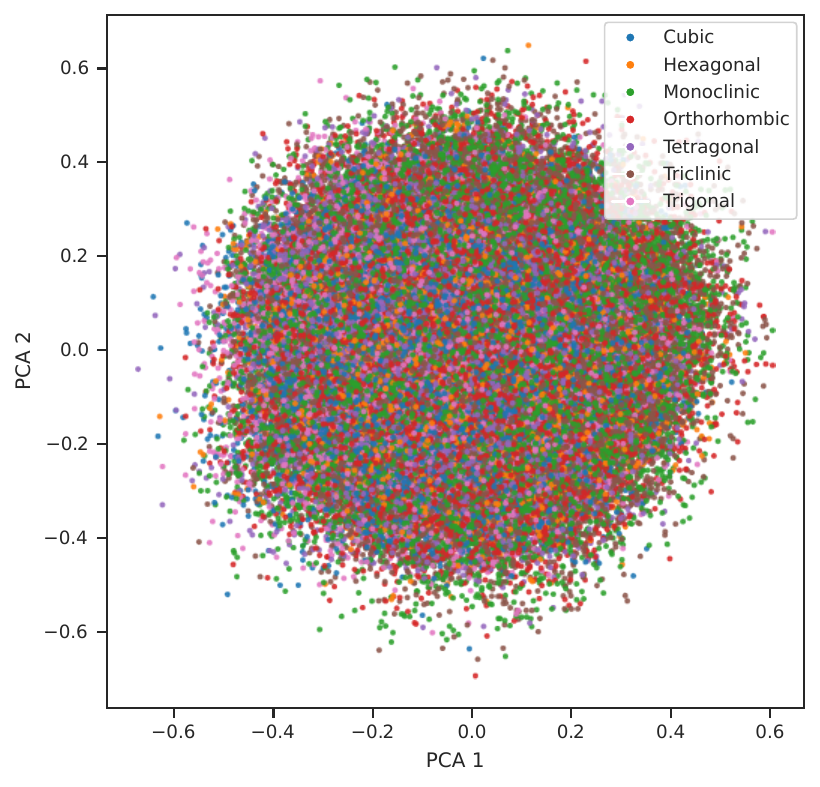}
         \caption{Colored by crystal system}
     \end{subfigure}
    \caption{\textbf{Interpretability of text embeddings visualized using PCA.}}
    \label{fig:embedding_pca}
\end{figure}

Figure~\ref{fig:embedding_umap} illustrates the embedding space after applying UMAP, which exhibits behavior similar to t-SNE.
We also apply PCA to further analyze the embedding space. As shown in Figure~\ref{fig:embedding_pca}, the first two principal components do not display any clear clustering structure. This outcome is expected for a contrastive learning framework, as it encourages the model to make use of all dimensions of the embedding space rather than concentrating information in just a few.

\section{Ablation studies}
\begin{figure}[h]
    \begin{subfigure}[t]{0.5\textwidth}
        \centering
        \includegraphics[width = \textwidth]{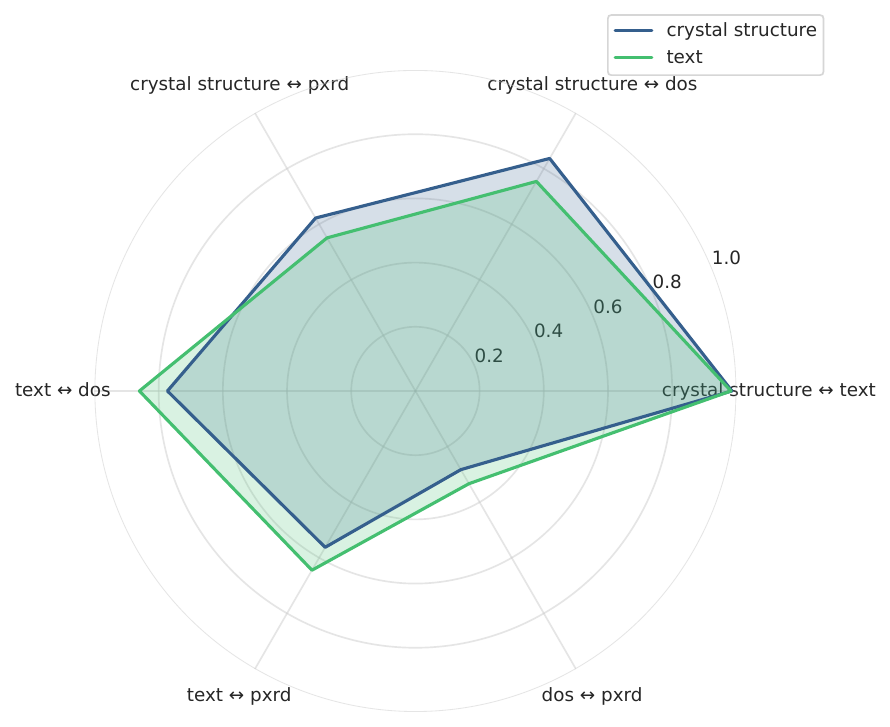}
        \caption{Cross-modal retrieval performance comparison between crystal structure and text modalities.}
    \label{fig:comparison_central_modality}
    \end{subfigure}
    \begin{subfigure}[t]{0.5\textwidth}
        \centering
        \includegraphics[width=\textwidth]{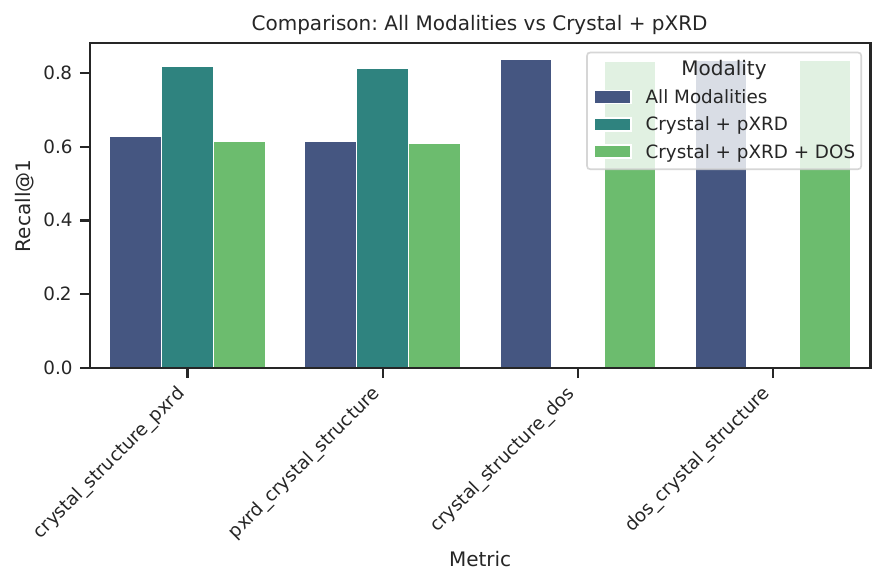}
        \caption{Effect of modality subset on crystal structure and \pxrd\ retrieval.}
        \label{fig:influence_different_modality}
    \end{subfigure}
    \caption{\textbf{Ablation study results.} (a) bidirectional retrieval performance for all modality pairs. (b) Recall@1 comparison between models trained on all four modalities versus only crystal structure and \pxrd.}
\end{figure}

We further investigate the impact of selecting different central modalities during training. Figure~\ref{fig:comparison_central_modality} compares recall performance when using text versus crystal structure as the central modality. No significant performance difference is observed between these two settings, suggesting that the learned alignment is robust to the choice of central modality.

In addition, we examine the influence of the number of modalities used during training. Three models are trained with an increasing number of modalities, and the results are shown in Figure~\ref{fig:influence_different_modality}. We observe that incorporating additional modalities leads to degraded alignment performance between crystal structure and \pxrd, while the alignment between crystal structure and \dos~ remains largely unaffected.

\section{Model Parameters, Training and Evaluation Settings}
\label{appdenix:model_para}

\paragraph{Model Parameters}
The four modality encoders and their key configurations are summarized in Table~\ref{tab:architectures}. Following each encoder, a projection head maps modality-specific embeddings into the shared alignment space. We adopt the projection head design from ImageBind~\cite{girdhar2023imagebind}, consisting of layer normalization followed by a linear layer that projects the encoder output to an alignment dimension of 128. For contrastive learning, a fixed temperature of $\tau = 0.07$ is used for all modality pairs, following CLIP~\cite{radford2021learning}. All training is performed using a fixed random seed to ensure reproducibility.
\begin{table}[h]
    \centering
    \caption{\textbf{Encoder architectures and parameter counts.}}
    \label{tab:architectures}
    \begin{tabular}{lllr}
        \hline
        Modality & Encoder & Key configuration & Parameters \\
        \hline
        Crystal structure & GCNN & 6 layers, latent dim 128, 
            41-component RBF & $\sim$488k \\
        \dos & Transformer & 2 layers, 8 heads, latent dim 512 & 
            $\sim$6.3M \\
        \pxrd & ResNet-1D & 62 Convolution blocks, latent dim 512 & 
            $\sim$120M \\
        Text & MatBERT & 12 layers, 12 attention heads, latent dim 768 & 
            $\sim$109M \\
        \hline
        \multicolumn{4}{l}{Projection head (all modalities): LayerNorm + 
            Linear, output dim 128} \\
        \hline
    \end{tabular}
\end{table}

\paragraph{\pxrd\ encoder architecture}
The \pxrd\ ResNet encoder processes one-dimensional intensity profiles of length 8501. The architecture follows~\cite{Schopmans_2023} and uses the ResNet 1D constructed from convolutional blocks. The model contains approximately 120M parameters. During contrastive training, the model is unfrozen.

\paragraph{MatBERT encoder architecture}

Textual descriptions are encoded using MatBERT~\cite{walker2021impact, Trewartha_2022}, a 12-layer bidirectional transformer with a hidden dimension of 768, 12 attention heads, and approximately 109M parameters. The final embedding is obtained by computing an average of token embeddings from the last hidden layer, excluding padding tokens. During contrastive training, the model is unfrozen unless explicitly stated.

\paragraph{Training Settings}
Due to the heterogeneous architectures and parameter scales of the encoders, employing a single optimizer and learning rate scheduler across all modalities leads to unstable training. Therefore, we use separate optimizers and schedulers for each encoder.

All encoders are trained using the AdamW optimizer with a base learning rate of $5 \times 10^{-4}$ and a weight decay of $1 \times 10^{-4}$. For the text encoder, the learning rate is scaled by a factor of 0.01 to account for its pretrained initialization.

For learning rate scheduling, the central modality encoder and the \dos\ encoder employ a cosine annealing schedule with warm-up. The remaining encoders use a ReduceLROnPlateau scheduler, where the validation loss with respect to the central modality is monitored.

\paragraph{Evaluation Setting}
We use recall as our primary evaluation metric, which reflects the fraction of queries for which the ground-truth paired sample appears within the top-k retrieved candidates. Retrieval is performed symmetrically in both directions between a modality pair, with the reported recall averaged over the two directions. Unless otherwise stated in the corresponding figure caption, $k=1$ and recall is averaged over both directions. The validation set serves as our retrieval candidate pool; the number of candidates for each modality is reported in Section~\ref{sec:dataset}. Since each query has exactly one ground-truth match, recall reduces to a simple fraction of correctly retrieved queries. As candidate pools are constructed directly from Materials Project entries, no ties or duplicate entries occur in the pool. For all retrieval evaluations, we use the projected embedding (i.e., the output after the projection head) rather than the backbone embedding. During evaluation, we use the Materials Project ID as the identifier for each material, ensuring that polymorphs of the same composition but different crystal systems are not conflated as a single identity.

\section{Sensitivity to the scoring weight $\alpha$}
\label{appendix:param_sensitivity}

To assess the robustness of the multimodal refinement procedure to the choice of $\alpha$, we evaluate Recall@1 and Recall@5 as a function of $\alpha \in [0, 1]$ for the two-stage \pxrd+text retrieval described in Section~\ref{sec:cases}.

\begin{figure}[h]
    \centering
    \includegraphics[width=0.5\linewidth]{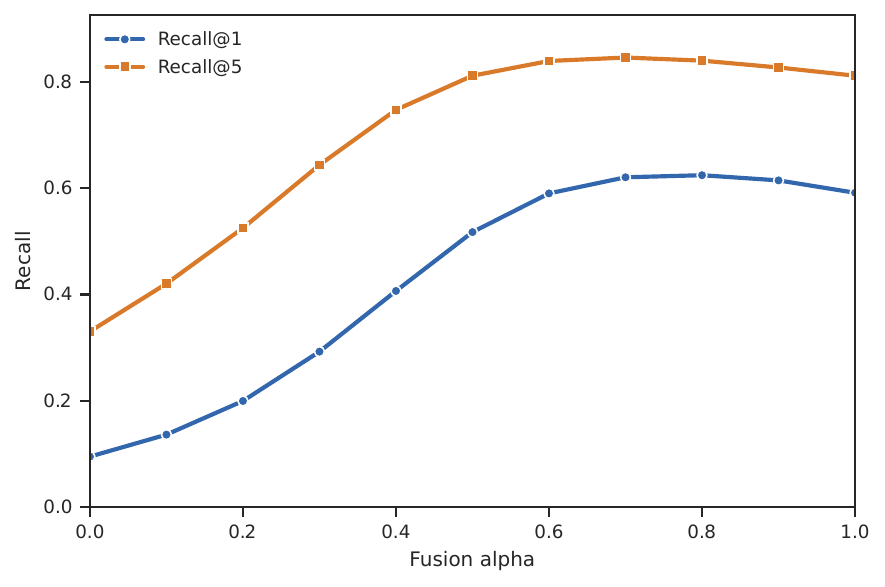}
    \caption{\textbf{Recall@1 and Recall@5 as a function of the fusion parameter $\alpha$}, which interpolates between the original embedding similarity ($\alpha=1$) and the refined retrieval with partial text ($\alpha=0$). Both metrics peak in the range $\alpha \in [0.7, 0.8]$, suggesting an optimal balance between the two scoring components. This indicates that introducing additional information can benefit retrieval and aligns well with real-world scenario.}
    \label{fig:placeholder}
\end{figure}

\section{Training convergence}

\begin{figure}[h]
    \centering
    \includegraphics[width=\linewidth]{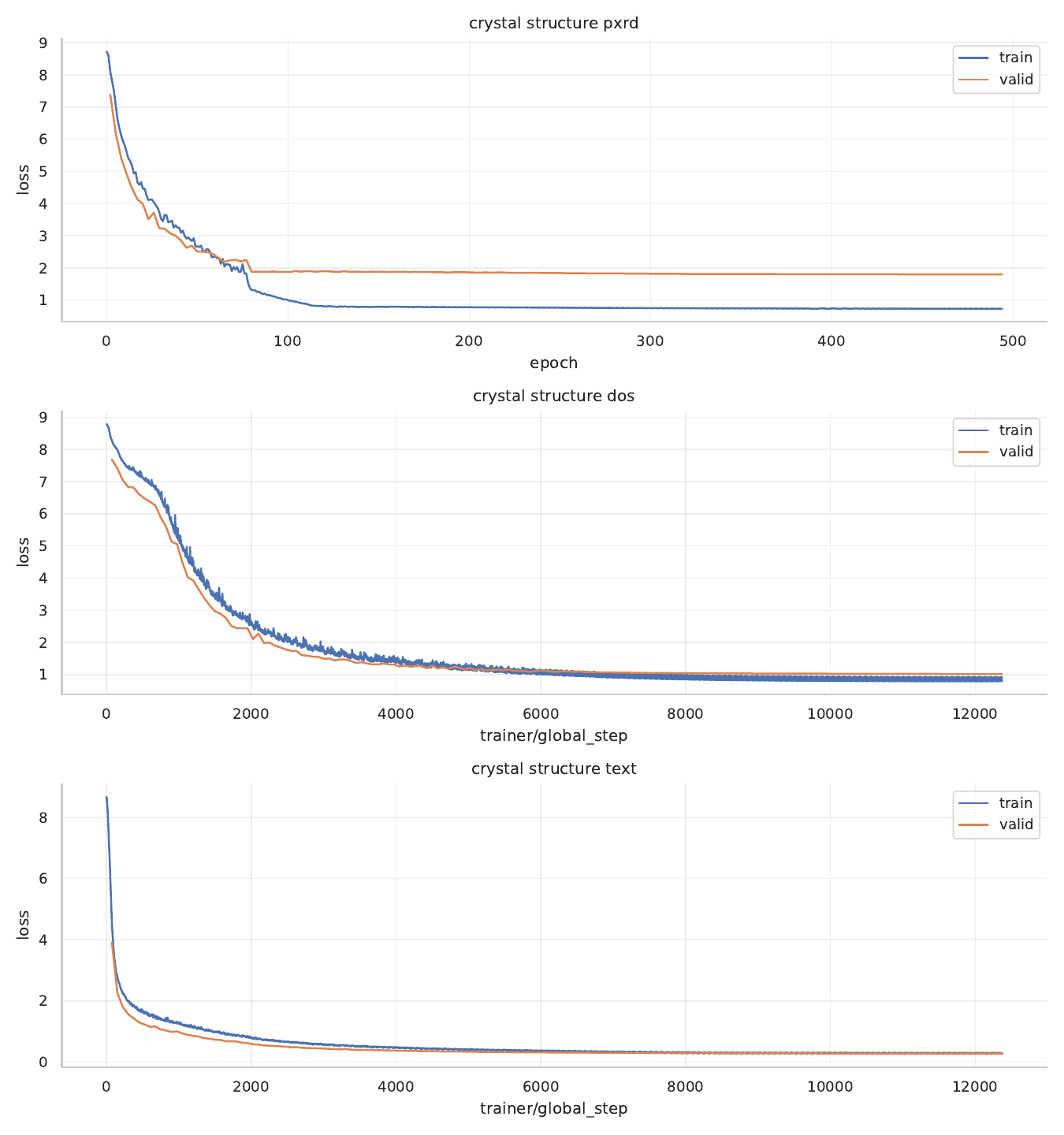}
    \caption{\textbf{Training convergence.} Contrastive loss curves for 
    each modality pair. Solid lines indicate training loss. Dashed lines indicate validation loss.}
    \label{fig:training_curves}
\end{figure}

Figure~\ref{fig:training_curves} shows the contrastive loss curves for each modality pair during training. Despite the heterogeneous encoder architectures and the use of separate optimizers and learning rate schedulers, all pairs converge in a stable manner.

\end{document}